\documentclass[10pt,twocolumn]{article}

\usepackage[utf8]{inputenc}
\usepackage[T1]{fontenc}
\usepackage{lmodern}
\usepackage{microtype}
\usepackage[letterpaper,margin=0.75in,columnsep=0.25in]{geometry}

\usepackage{amsmath,amssymb,amsfonts,mathtools,bm}
\usepackage{graphicx}
\usepackage{booktabs}
\usepackage{multirow}
\usepackage{array}
\usepackage{tabularx}
\usepackage{algorithm}
\usepackage{algpseudocode}
\usepackage{enumitem}
\usepackage{xcolor}
\usepackage[font=small,labelfont=bf]{caption}
\usepackage[numbers,sort&compress]{natbib}
\usepackage[colorlinks=true,allcolors=black]{hyperref}
\hypersetup{
  pdftitle={GeoRouteNet: A Geometry-Aware Non-Autoregressive Neural Solver for the Euclidean Traveling Salesman Problem},
  pdfauthor={Xiang Li}
}

\setlist{leftmargin=*,nosep}

\newcommand{\RR}{\mathbb{R}}
\newcommand{\EE}{\mathbb{E}}

\newcommand{\method}{GeoRouteNet}
\newcommand{\mcsrl}{MCS-RL}
\newcommand{\nar}{NAR4TSP}
\newcommand{\narR}{\nar--PG}
\newcommand{\narMCS}{\nar--\mcsrl}
\newcommand{\grnPG}{\method--PG}
\newcommand{\grnMCS}{\method--\mcsrl}
\newcommand{\lkh}{LKH-3}

\newcommand{\tspfifty}{TSP-50}
\newcommand{\tsponehundred}{TSP-100}
\newcommand{\tsplib}{TSPLIB}

\newcommand{\nodeset}{V}
\newcommand{\numnodes}{n}
\newcommand{\coords}{\bm{x}}
\newcommand{\centered}{\bm{c}}
\newcommand{\dist}{d}
\newcommand{\tourperm}{\pi}
\newcommand{\tourcost}{C}

\newcommand{\rbfcount}{K_{\mathrm{rbf}}}
\newcommand{\numcandidates}{M}
\newcommand{\baseline}{b}
\newcommand{\advantage}{A}
\newcommand{\normadvantage}{\widehat{A}}
\newcommand{\entropycoef}{\eta}
\newcommand{\guidancecoef}{\alpha}

\DeclareMathOperator{\softmax}{softmax}
\DeclareMathOperator{\SiLU}{SiLU}
\DeclareMathOperator{\LayerNorm}{LayerNorm}

\title{\method: A Geometry-Aware Non-Autoregressive Neural Solver\\
for the Euclidean Traveling Salesman Problem}

\author{Xiang Li\\[2pt]
\normalsize College of Computer Science, Yangtze University, Jingzhou, China\\
\normalsize \texttt{lixiang.stu@yangtzeu.edu.cn}}
\date{}

\begin{document}
\maketitle

\begin{abstract}
Non-autoregressive neural solvers amortize computation across traveling salesman problem (TSP) instances, but models trained on random Euclidean instances can degrade when the number or spatial distribution of nodes changes.
We study whether explicit geometric features and a richer within-instance training signal improve transfer across graph sizes and spatial distributions.
We introduce \method, which augments a non-autoregressive TSP solver with centered node offsets and radii, learnable radial distance bases, distance-aware graph attention, explicit edge messages, and cross-layer representation mixing.
We also introduce multi-candidate self-comparison reinforcement learning (\mcsrl), which trains on several sampled tours per instance using a leave-one-out adaptive baseline, winner-candidate guidance, and annealed entropy regularization.
In a single-seed study, all neural variants are trained only on random \tspfifty\ instances and evaluated with the same greedy and beam-search decoders.
Under Beam-1000 decoding, \grnMCS{} obtains gaps of $0.32\%$ on the \tspfifty\ validation set used for checkpoint selection, $1.26\%$ on a correlated \tsponehundred\ size diagnostic, and $3.60\%$ across 27 \tsplib\ EUC\_2D instances.
The \narR{} gaps on the same evaluations are $0.42\%$, $2.73\%$, and $17.12\%$.
A $2\!\times\!2$ comparison crosses encoder and training choices.  Under PG, the geometry-aware encoder has lower gaps than the reproduced encoder on the \tsponehundred\ diagnostic and \tsplib.  With the geometry-aware encoder, \mcsrl\ is associated with a further reduction; with the reproduced encoder, it has a higher \tsplib\ gap.
\end{abstract}

\section{Introduction}
\label{sec:introduction}

The traveling salesman problem (TSP) asks for the shortest Hamiltonian cycle through a set of nodes.
It is a canonical combinatorial optimization problem with mature exact and heuristic solvers, including Concorde and the Lin--Kernighan--Helsgaun family~\cite{applegate2006tsp,lin1973effective,helsgaun2000effective,helsgaun2017extension}.
These solvers provide strong solution quality through instance-specific search.
Neural combinatorial optimization instead learns a reusable mapping from instances to solutions, making it attractive when many related instances must be processed under a common computational budget~\cite{vinyals2015pointer,bello2016neural,kool2019attention}.

Most early neural TSP solvers are autoregressive: they select one node at a time while conditioning on the partial tour~\cite{vinyals2015pointer,kool2019attention,kwon2020pomo}.
This factorization is expressive but requires a sequence of neural-network evaluations.
Non-autoregressive edge-prediction approaches compute pairwise scores in parallel and impose feasibility during decoding~\cite{joshi2019efficient,xiao2025reinforcement}.
Diffusion solvers provide another non-autoregressive formulation through iterative denoising~\cite{sun2023difusco,wang2025efficient}.
Generalization across graph sizes and spatial distributions nevertheless remains difficult.
A model trained on uniformly sampled points of one size can rely on coordinate ranges, local densities, or graph statistics that do not transfer to larger or structurally different instances~\cite{joshi2022learning}.

We take \nar~\cite{xiao2025reinforcement} as the reference non-autoregressive framework.
In our single-seed reproduction, its Beam-1000 gap is $0.42\%$ on the \tspfifty\ validation set used for checkpoint selection, $2.73\%$ on a correlated \tsponehundred\ size diagnostic, and $17.12\%$ over 27 \tsplib\ EUC\_2D instances.
These measurements motivate two questions.
First, can the encoder expose Euclidean structure more directly instead of requiring it to recover all geometric relationships from raw coordinates?
Second, can several tours sampled from the same policy provide a more informative training comparison than one sampled tour and one greedy baseline?
We treat these as empirical questions and evaluate the two interventions both separately and together.

Our model, \method, retains the start-node and edge-transition outputs of \nar\ while replacing its encoder with a geometry-aware alternative.
The node input augments raw coordinates with centroid-relative offsets and radial position.
Pairwise distances are expanded through learnable radial basis functions and then used in both edge states and attention logits.
The encoder jointly updates node and edge representations, and combines shallow and deep node states through attention over layer history.
These choices inject distance and relative-position information without changing the downstream tour decoder.

We complement the encoder with multi-candidate self-comparison reinforcement learning (\mcsrl).
For each instance, \mcsrl\ samples several tours and compares each candidate with the greedy tour and the best of its peer candidates.
A separate winner term increases the likelihood of the shortest sampled candidate in proportion to its advantage over the candidate mean, while an entropy bonus is annealed over optimization steps.
The method changes training only; evaluation uses the same greedy, Beam-100, or Beam-1000 decoder for all neural variants.

The experiments cross two encoder choices (reproduced \nar\ or geometry-aware) with two training objectives (single-candidate policy gradient or \mcsrl), yielding \narR, \narMCS, \grnPG, and \grnMCS.
The last combination is the full \method\ configuration.
All variants are trained on random \tspfifty\ with one training seed.
With policy gradient fixed, replacing the encoder changes the Beam-1000 gap from $2.73\%$ to $1.54\%$ on \tsponehundred\ and from $17.12\%$ to $5.41\%$ on \tsplib.
Pairing \mcsrl\ with the geometry-aware encoder changes the gaps further to $1.26\%$ and $3.60\%$, respectively.
In contrast, pairing \mcsrl\ with the reproduced \nar\ encoder gives a \tsplib\ gap of $21.98\%$.
The evidence therefore supports only a qualified descriptive conclusion: under PG, replacing the reproduced encoder yields a larger numerical gap reduction on both transfer evaluations than switching from PG to \mcsrl\ within the geometry-aware encoder, while the direction of the training-objective contrast differs across encoders.

Our contributions are threefold:
\begin{itemize}
    \item We develop a geometry-aware non-autoregressive encoder that integrates absolute and centroid-relative node channels, learnable distance bases, distance-conditioned attention, edge-state updates, and cross-layer representation mixing.
    \item We formulate \mcsrl, a within-instance multi-sample objective combining a leave-one-out adaptive baseline, winner-candidate guidance, and entropy annealing.
    \item We report a single-seed $2\!\times\!2$ comparison on a \tspfifty\ validation set, a correlated \tsponehundred\ size diagnostic, and 27 \tsplib\ instances, documenting the encoder-dependent pattern without treating it as a component-level causal attribution.
\end{itemize}

Section~\ref{sec:related_work} positions the study within neural combinatorial optimization.
The remaining sections define the problem, present \method\ and \mcsrl, describe the evaluation protocol, and discuss the scope and limitations of the results.

\section{Related Work}
\label{sec:related_work}

\paragraph{Autoregressive neural combinatorial optimization.}
Pointer Networks established sequence-to-sequence construction for combinatorial problems~\cite{vinyals2015pointer}, and policy-gradient training made it possible to optimize tour length without labeled optimal tours~\cite{bello2016neural,williams1992simple}.
Attention-based routing models subsequently improved the encoder and rollout baseline~\cite{kool2019attention}, while POMO exploited multiple equivalent starts to represent the multiplicity of high-quality solutions~\cite{kwon2020pomo}.
Learned construction has been extended to vehicle routing~\cite{nazari2018reinforcement}, while neural improvement methods iteratively refine routing solutions~\cite{hottung2020neural,ma2021learning}.
These methods obtain strong solutions, but autoregressive construction evaluates a neural decision rule repeatedly along the tour.
Our method instead uses a non-autoregressive encoder and static policy heads, with sequential computation confined to the feasibility decoder.

\paragraph{Non-autoregressive and diffusion-based solvers.}
Graph-convolutional TSP models predict edge scores in parallel and recover a tour through search or decoding~\cite{joshi2019efficient}.
DIMES learns non-autoregressive solution distributions that guide combinatorial search~\cite{wu2022dimes}.
Diffusion-based methods instead learn an iterative denoising process over solution representations~\cite{sun2023difusco,wang2025efficient}.
\nar\ directly predicts a start distribution and an edge-transition distribution and trains these outputs with reinforcement learning~\cite{xiao2025reinforcement}.
This formulation is the closest reference for our study: \method\ preserves its output factorization and decoding interface, but changes the geometric encoder, while \mcsrl\ changes the training objective.
Recent work on scalable neural TSP solving and neural-guided classical search further illustrates that representation quality and post-processing are complementary design choices~\cite{wen2025localescaper,xin2021neurolkh}.

\paragraph{Geometry and reinforcement-learning signals.}
Graph-attention and self-attention layers can serve as permutation-equivariant node aggregators when no sequence positional encoding is used; without geometry-specific inputs or biases, they do not explicitly represent Euclidean distances or coordinate transformations~\cite{velickovic2018graph,vaswani2017attention}.
Sym-NCO exploits known problem symmetries through symmetry-aware training~\cite{kim2022symnco}; \method\ instead injects centroid-relative coordinates and pairwise-distance features directly into the encoder.
Empirical studies of neural TSP solvers show that performance on one random-instance distribution does not imply transfer across graph sizes or spatial distributions~\cite{joshi2022learning}.
This evidence motivates the explicit relative coordinates, radial distance features, and distance-conditioned messages used here.
On the optimization side, REINFORCE uses baselines to reduce gradient variance~\cite{williams1992simple,sutton2018reinforcement}; routing methods commonly use learned, rollout, greedy, or multi-start comparisons~\cite{kool2019attention,kwon2020pomo,xiao2025reinforcement}.
\mcsrl\ is related to these multi-sample ideas but uses a within-instance leave-one-out baseline together with a separate winner-guidance term.
Our factorial comparison evaluates the training signal with both encoder choices rather than attributing all observed differences to a single combined system.

\section{Problem Formulation}
\label{sec:problem}

\subsection{Euclidean TSP}
An instance is a complete graph $G=(\nodeset,E)$ with
$\nodeset=\{1,\ldots,\numnodes\}$ and planar node coordinates
$\coords_i\in\RR^2$.  Its edge costs are Euclidean distances
\begin{equation}
    \dist_{ij}=\lVert \coords_i-\coords_j\rVert_2.
    \label{eq:distance}
\end{equation}
A feasible tour is a permutation $\tourperm=(\tourperm_1,\ldots,
\tourperm_{\numnodes})$ of all nodes.  Closing the cycle with
$\tourperm_{\numnodes+1}=\tourperm_1$, its cost is
\begin{equation}
    \tourcost(\tourperm;\coords)
    =\sum_{t=1}^{\numnodes}
      \dist_{\tourperm_t,\tourperm_{t+1}}.
    \label{eq:tour_cost}
\end{equation}
The objective is to find
$\tourperm^*\in\arg\min_{\tourperm\in S_{\numnodes}}
\tourcost(\tourperm;\coords)$.  Training instances in this work have
coordinates sampled independently and uniformly from $[0,1]^2$.

\subsection{Static-Score Neural Policy}
Following the non-autoregressive edge-prediction formulation of
\nar~\citep{xiao2025reinforcement}, the neural network evaluates an
instance once and returns (i) a start-node distribution
$p_\theta^{\mathrm{s}}(i\mid G)$ and (ii) a matrix of directed edge logits
$z_{ij}$.  Let $U_t=\nodeset\setminus
\{\tourperm_1,\ldots,\tourperm_t\}$ denote the unvisited nodes after
step $t$.  The decoder converts the static logits into a masked
transition distribution,
\begin{equation}
    p_\theta^{\mathrm{e}}(j\mid i,U_t,G)
    =\frac{\exp(z_{ij})}
    {\sum_{u\in U_t}\exp(z_{iu})},
    \qquad j\in U_t,
    \label{eq:masked_transition}
\end{equation}
and assigns a tour the probability
\begin{equation}
    p_\theta(\tourperm\mid G)
    =p_\theta^{\mathrm{s}}(\tourperm_1\mid G)
     \prod_{t=1}^{\numnodes-1}
     p_\theta^{\mathrm{e}}(
        \tourperm_{t+1}\mid\tourperm_t,U_t,G).
    \label{eq:tour_factorization}
\end{equation}
The return edge $(\tourperm_{\numnodes},\tourperm_1)$ contributes to
the cost in Eq.~\eqref{eq:tour_cost}, but it is fixed once the
permutation has been constructed and is therefore not a separate
policy action.
Thus, ``non-autoregressive'' refers to computing all neural node and
edge scores in one encoder pass.  Feasibility is imposed by a
lightweight sequential decoder whose visited-node mask changes, but
whose neural representations and edge logits are not recomputed.
Without optimal-tour supervision, the task-level objective is to
reduce the expected tour cost
\begin{equation}
    \mathcal{J}(\theta)
    =\EE_{G}\EE_{\tourperm\sim p_\theta(\cdot\mid G)}
      [\tourcost(\tourperm;\coords)].
    \label{eq:expected_cost}
\end{equation}
We optimize policy-gradient surrogates of this objective.
Section~\ref{sec:mcsrl} describes both the single-candidate reference
objective and \mcsrl, which adds winner guidance and entropy
regularization.

\section{GeoRouteNet}
\label{sec:method}

\begin{figure*}[t]
    \centering
    \includegraphics[width=0.94\textwidth]{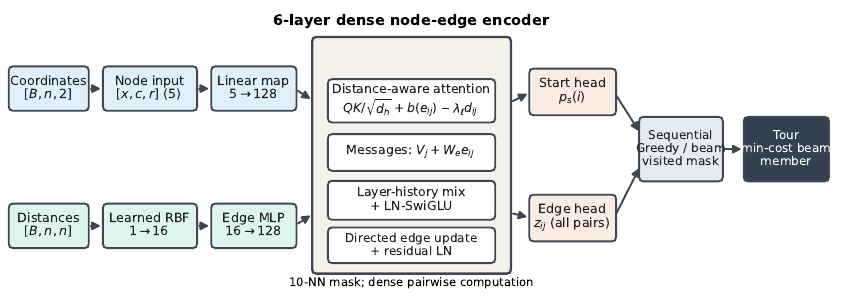}
    \caption{Architecture of \method.  Dense node-pair distances are
    encoded by learnable radial basis functions, while a 10-nearest-
    neighbor mask restricts which pairs participate in node attention.
    The encoder jointly updates node and directed edge states, mixes
    node representations across depth, and produces a start-node
    distribution and dense directed edge logits.  The mask is an
    inductive bias rather than a sparse-computation mechanism in the
    reported formulation.}
    \label{fig:architecture}
\end{figure*}

\method{} retains the static start-and-edge policy in
Section~\ref{sec:problem}, while redesigning its geometric input
encoding and node--edge message passing.  Figure~\ref{fig:architecture}
summarizes the computation.  The reported configuration uses hidden dimension
$d=128$, six encoder layers, eight graph-attention heads, and a
two-layer edge decoder.

\subsection{Geometric Input Encoding}

\paragraph{Node channels.}
For an instance centroid
$\bar{\coords}=\numnodes^{-1}\sum_j\coords_j$, we define
\begin{equation}
    \centered_i=\coords_i-\bar{\coords},
    \qquad r_i=\lVert\centered_i\rVert_2,
    \qquad
    \bm{u}_i=[\coords_i\,\Vert\,\centered_i\,\Vert\,r_i].
    \label{eq:node_features}
\end{equation}
The initial node state is $\bm{h}_i^{(0)}=W_x\bm{u}_i+b_x$.
The centered and radial channels are translation invariant and expose
instance-relative geometry.  The complete input is \emph{not}
strictly translation invariant, however, because it deliberately
retains the original coordinates $\coords_i$ as two additional
channels.  We therefore describe Eq.~\eqref{eq:node_features} as a
geometric augmentation rather than an invariance guarantee.

\paragraph{Dense distances and the neighborhood mask.}
We first compute all $\numnodes^2$ pairwise distances.  For node $i$,
$\mathcal{N}_k(i)$ contains $i$ itself and its $k=10$ nearest
non-self nodes; distance ties at the boundary are retained.  The
binary mask $A_{ij}=\mathbf{1}[j\in\mathcal{N}_k(i)]$ is used only in
node attention.  Edge features and decoded logits are maintained for
every ordered pair $(i,j)$, including pairs outside this mask.

\paragraph{Learnable radial basis functions.}
Scalar distances are lifted with $\rbfcount=16$ Gaussian bases,
\begin{equation}
    \phi_k(\dist_{ij})
    =\exp\!\left[-\left(
      \frac{\dist_{ij}-\mu_k}{\sigma}
    \right)^2\right],
    \qquad k=1,\ldots,\rbfcount.
    \label{eq:rbf}
\end{equation}
Both the centers and their shared positive bandwidth are learned.
Specifically, unconstrained parameters $a_k$ and $\rho$ are mapped as
\begin{equation}
    \bm{\mu}
      =\operatorname{sort}\!\left(
        \sqrt{2}\,\operatorname{sigmoid}(\bm{a})
      \right),
    \qquad
    \sigma=\max\{\exp(\rho),10^{-3}\}.
    \label{eq:rbf_parameterization}
\end{equation}
The centers are initialized from a uniform grid on $[0,\sqrt{2}]$;
the endpoint fractions are clamped to $10^{-4}$ and $1-10^{-4}$
before conversion to logits.  The shared $\sigma$ is initialized to
$\sqrt{2}/(\rbfcount-1)$.  For the reported
configuration, no raw $\dist_{ij}$ or $\dist_{ij}^2$ channel is
concatenated to the RBF vector.  A two-layer SiLU MLP produces the
initial dense edge state
\begin{equation}
    \bm{e}_{ij}^{(0)}
    =W_{e,2}\SiLU(W_{e,1}\bm{\phi}(\dist_{ij})+\bm{b}_{e,1})
      +\bm{b}_{e,2}.
    \label{eq:edge_embedding}
\end{equation}

\subsection{Distance-Aware Node--Edge Encoder}

Let $\bm{h}_i^{(\ell-1)}$ and
$\bm{e}_{ij}^{(\ell-1)}$ be the node and directed edge states entering
layer $\ell$.  For attention head $q$, linear projections produce
$\bm{q}_i^{\ell q}$, $\bm{k}_j^{\ell q}$, and
$\bm{v}_j^{\ell q}$.  The current edge state supplies both a
head-specific logit bias $b_{ij}^{\ell q}$ and an edge value
$\bm{v}_{e,ij}^{\ell q}$.  Attention is
\begin{align}
    s_{ij}^{\ell q}
      &=\frac{\langle\bm{q}_i^{\ell q},
                        \bm{k}_j^{\ell q}\rangle}{\sqrt{d_q}}
        +b_{ij}^{\ell q}
        -\lambda_\ell\dist_{ij},
        \qquad
        \lambda_\ell=\operatorname{softplus}(\gamma_\ell),
        \label{eq:attention_logit}\\
    \alpha_{ij}^{\ell q}
      &=\frac{\exp(s_{ij}^{\ell q})}
        {\sum_{u\in\mathcal{N}_k(i)}\exp(s_{iu}^{\ell q})},
        \qquad j\in\mathcal{N}_k(i),
        \label{eq:attention_weights}\\
    \bm{g}_i^{(\ell)}
      &=W_o^\ell\mathop{\Vert}_{q=1}^{H}
        \sum_{j\in\mathcal{N}_k(i)}
        \alpha_{ij}^{\ell q}
        (\bm{v}_j^{\ell q}+\bm{v}_{e,ij}^{\ell q}).
        \label{eq:node_message}
\end{align}
There is one learned distance coefficient $\lambda_\ell$ per encoder
layer, shared by the heads within that layer.  This is distinct from a
single coefficient shared across the complete network.

\paragraph{Cross-layer attentive residual.}
At layer $\ell$, a residual mixer attends over the node's history
$\mathcal{H}_i^{\ell-1}=\{\bm{h}_i^{(0)},\ldots,
\bm{h}_i^{(\ell-1)}\}$.  With projected query
$W_{Q,\ell}^R\bm{g}_i^{(\ell)}$, projected keys
$W_{K,\ell}^R\bm{h}_i^{(r)}$, and the
unprojected historical states as values, it computes
\begin{align}
    \beta_{ir}^{(\ell)}
      &=\softmax_{r=0,\ldots,\ell-1}
        \left(
        \frac{\langle W_{Q,\ell}^R\bm{g}_i^{(\ell)},
                       W_{K,\ell}^R\bm{h}_i^{(r)}\rangle}{\sqrt{d_R}}
        \right),
        \label{eq:residual_attention}\\
    \bm{r}_i^{(\ell)}
      &=\sum_{r=0}^{\ell-1}\beta_{ir}^{(\ell)}\bm{h}_i^{(r)},
    \qquad
    \widetilde{\bm{h}}_i^{(\ell)}
      =\LayerNorm(\bm{g}_i^{(\ell)}+\bm{r}_i^{(\ell)}).
    \label{eq:residual_mix}
\end{align}
The reported model uses one residual-attention head.  A SwiGLU
feed-forward block~\citep{shazeer2020glu} with expansion factor two
then gives
\begin{align}
    \operatorname{FFN}(\bm{h})
      &=W_f\bigl[\SiLU(W_g\bm{h}+\bm{b}_g)
        \odot(W_v\bm{h}+\bm{b}_v)\bigr]+\bm{b}_f,
        \label{eq:swiglu}\\
    \bm{h}_i^{(\ell)}
      &=\LayerNorm\!\left(
        \widetilde{\bm{h}}_i^{(\ell)}
        +\operatorname{FFN}(\widetilde{\bm{h}}_i^{(\ell)})
      \right).
    \label{eq:node_ffn}
\end{align}
Layer normalization~\citep{ba2016layer} is applied over each node's
feature dimension and does not depend on batch statistics.

After the node update, every directed edge state is refined as
\begin{align}
    \bm{u}_{ij}^{(\ell)}
      &=W_e^\ell\bm{e}_{ij}^{(\ell-1)}
        +W_s^\ell\bm{h}_i^{(\ell)}
        +W_t^\ell\bm{h}_j^{(\ell)}+\bm{b}_e^\ell,
        \label{eq:edge_proposal}\\
    \bm{e}_{ij}^{(\ell)}
      &=\LayerNorm\!\left(
        \bm{e}_{ij}^{(\ell-1)}
        +\SiLU(\bm{u}_{ij}^{(\ell)})
      \right).
      \label{eq:edge_update}
\end{align}
The source and target projections are distinct, and the reported
configuration does not symmetrize edge states.

\subsection{Policy Heads and Decoding}

A learned start query attends to the node states after every encoder
layer.  Its query representation is updated with residual layer
normalization through the first five layers; the head-averaged
attention weights at the final layer form $p_\theta^{\mathrm{s}}$.
For each ordered pair, a two-layer ReLU MLP maps
$\bm{e}_{ij}^{(6)}$ to the scalar logit $z_{ij}$.  The decoder then
applies Eq.~\eqref{eq:masked_transition} over \emph{all} unvisited
nodes, not only the 10-nearest-neighbor attention set.  Greedy decoding
chooses the maximum-probability feasible action.  Beam decoding keeps
partial tours by cumulative log-probability and, after completion,
returns the minimum-cost tour among the retained full candidates.

\subsection{Computational Complexity}

Our implementation is dense.  It materializes pairwise
distances, RBF edge features, $\numnodes\times\numnodes$ edge states,
and full attention logits before applying the neighborhood mask.
Consequently, the encoder remains quadratic in $\numnodes$: its dense
edge projections require
$O(BL\numnodes^2d^2)$ arithmetic in a direct accounting, while its
dominant pairwise activation memory is $O(B\numnodes^2d)$.
The 10-nearest-neighbor mask changes information flow but does not
reduce these bounds to $O(B\numnodes k)$ without a sparse kernel.
Greedy decoding costs $O(B\numnodes^2)$; beam search with width $W$
costs $O(BW\numnodes^2)$ for sequential expansion, in addition to the
single encoder pass.

\section{MCS-RL Training}
\label{sec:mcsrl}

The network is trained without optimal tours.  REINFORCE estimates
policy gradients from sampled trajectories~\citep{williams1992simple}.
The single-sample objective in our \nar\ reproduction compares one sampled
tour with greedy decoding from the same model~\citep{xiao2025reinforcement},
but a single draw exposes only one realization of the policy on each
instance.  We introduce \textbf{multi-candidate self-comparison
reinforcement learning} (\mcsrl), which draws several tours from the
same static-score policy and constructs candidate-specific learning
signals from their relative costs.  The method changes only training;
the policy heads and test-time decoders remain those in
Section~\ref{sec:method}.

Algorithm~\ref{alg:mcsrl} summarizes one optimizer step.  The
subsections below define its candidate-specific baselines, winner
guidance, and path-conditional entropy terms.

\begin{algorithm}[H]
\caption{One \mcsrl{} optimizer step.}
\label{alg:mcsrl}
\small
\begin{algorithmic}[1]
\Require minibatch $\mathcal{B}=\{G_b\}_{b=1}^{B}$ and parameters $\theta$
\Require $\numcandidates\geq2$, $\guidancecoef$, $\entropycoef_0$, $\entropycoef_T$, step $t$, and total steps $T$
\State $(p_b^{\mathrm{s}},z_b)_{b=1}^{B}\gets f_\theta(\mathcal{B})$
\For{$b=1,\ldots,B$}
    \For{$m=1,\ldots,\numcandidates$}
        \State $(\tourperm_b^{(m)},\ell_b^{(m)},H_b^{(m)})\gets
        \Call{SampleDecode}{p_b^{\mathrm{s}},z_b}$
        \State $C_b^{(m)}\gets \tourcost(\tourperm_b^{(m)};\coords_b)$
    \EndFor
    \State $\tourperm_b^{\mathrm{g}}\gets
    \Call{GreedyDecode}{p_b^{\mathrm{s}},z_b}$
    \State $C_b^{\mathrm{g}}\gets
    \tourcost(\tourperm_b^{\mathrm{g}};\coords_b)$
    \For{$m=1,\ldots,\numcandidates$}
        \State $\baseline_b^{(m)}\gets
        \min\{C_b^{\mathrm{g}},\min_{r\ne m}C_b^{(r)}\}$
        \State $\advantage_b^{(m)}\gets
        C_b^{(m)}-\baseline_b^{(m)}$
    \EndFor
    \State $m_b^*\gets\arg\min_m C_b^{(m)}$
    \State $\Delta_b\gets
    \numcandidates^{-1}\sum_m C_b^{(m)}-C_b^{(m_b^*)}$
\EndFor
\State $\{\normadvantage_b^{(m)}\}_{b,m}\gets
\Call{Standardize}{\{\advantage_b^{(m)}\}_{b,m}}$
\State Compute $\mathcal{L}_{\mathrm{PG}}$ by Eq.~\eqref{eq:mcs_policy_gradient}
\State Compute $\mathcal{L}_{\mathrm{lead}}$ and $\bar H$ by
Eqs.~\eqref{eq:winner_loss} and~\eqref{eq:mean_path_entropy}
\State $\entropycoef_t\gets
\entropycoef_0+(\entropycoef_T-\entropycoef_0)t/(T-1)$
\State $\mathcal{L}\gets
\mathcal{L}_{\mathrm{PG}}+\guidancecoef\mathcal{L}_{\mathrm{lead}}
-\entropycoef_t\bar H$
\State $\theta\gets\Call{Adam}{\theta,\nabla_\theta\mathcal{L}}$
\end{algorithmic}
\end{algorithm}

\subsection{Candidate Sampling and Adaptive Baselines}

Consider a minibatch of $B$ instances and $\numcandidates$ sampled
tours per instance.  For instance $b$ and candidate $m$, we sample
$\tourperm_b^{(m)}\sim p_\theta(\cdot\mid G_b)$ and set
$C_b^{(m)}=\tourcost(\tourperm_b^{(m)};\coords_b)$.  Define
$p_{b,0}=p_\theta^{\mathrm{s}}(\cdot\mid G_b)$.  For $t\geq1$,
$p_{b,t}^{(m)}$ is the masked distribution in
Eq.~\eqref{eq:masked_transition} after the first $t$ nodes of
candidate $m$ have been sampled.  Its full
trajectory log-probability is
\begin{equation}
    \ell_b^{(m)}
    =\log p_{b,0}(\tourperm_{b,1}^{(m)})
     +\sum_{t=1}^{\numnodes-1}
      \log p_{b,t}^{(m)}(\tourperm_{b,t+1}^{(m)})
    \label{eq:candidate_logprob}
\end{equation}
We also greedily decode the
same model outputs and denote the resulting cost by $C_b^{\mathrm{g}}$.
For candidate $m$, the adaptive leave-one-candidate-out baseline is
\begin{equation}
    \baseline_b^{(m)}
    =\min\!\left\{
      C_b^{\mathrm{g}},
      \min_{r\ne m}C_b^{(r)}
    \right\},
    \qquad
    \advantage_b^{(m)}=C_b^{(m)}-\baseline_b^{(m)}.
    \label{eq:adaptive_baseline}
\end{equation}
For the degenerate case $\numcandidates=1$, we set
$\baseline_b^{(1)}=C_b^{\mathrm{g}}$.

We standardize advantages jointly over all
$B\numcandidates$ candidates in the minibatch, rather than separately
within each instance.  Define
\begin{align}
    \bar A
      &=\frac{1}{B\numcandidates}
        \sum_{b=1}^{B}\sum_{m=1}^{\numcandidates}
        \advantage_b^{(m)},
        \label{eq:advantage_mean}\\
    s_A
      &=\left[
        \frac{1}{B\numcandidates}
        \sum_{b=1}^{B}\sum_{m=1}^{\numcandidates}
        (\advantage_b^{(m)}-\bar A)^2
        \right]^{1/2},
        \label{eq:advantage_std}\\
    \normadvantage_b^{(m)}
      &=\frac{\advantage_b^{(m)}-\bar A}
      {\max\{s_A,10^{-6}\}}.
    \label{eq:advantage_standardization}
\end{align}
The standardized policy-gradient term is
\begin{equation}
    \mathcal{L}_{\mathrm{PG}}
    =\frac{1}{B\numcandidates}
      \sum_{b=1}^{B}\sum_{m=1}^{\numcandidates}
      \operatorname{sg}\!\left[
        \normadvantage_b^{(m)}
      \right]\ell_b^{(m)},
    \label{eq:mcs_policy_gradient}
\end{equation}
where $\operatorname{sg}$ stops gradients.  Under gradient descent,
a candidate that is more costly than its comparison baseline has a
positive advantage and its log-probability is reduced.

\subsection{Winner-Candidate Guidance}

Let $m_b^*\in\arg\min_m C_b^{(m)}$ be the shortest sampled candidate
for instance $b$, and define its improvement over the candidate mean
as
\begin{equation}
    \Delta_b
    =\frac{1}{\numcandidates}\sum_{m=1}^{\numcandidates}C_b^{(m)}
      -C_b^{(m_b^*)}.
    \label{eq:winner_gain}
\end{equation}
The winner-guidance term increases the likelihood of this candidate
in proportion to that stop-gradient cost difference:
\begin{equation}
    \mathcal{L}_{\mathrm{lead}}
    =-\frac{1}{B}\sum_{b=1}^{B}
      \operatorname{sg}[\Delta_b]
      \frac{\ell_b^{(m_b^*)}}{\numnodes}.
    \label{eq:winner_loss}
\end{equation}
Dividing the accumulated trajectory log-probability by $\numnodes$
puts this auxiliary term on a per-decision scale.  The construction is
within-instance: candidate costs from different TSP instances are
never compared in Eq.~\eqref{eq:winner_loss}.

\subsection{Path-Conditional Entropy and Full Objective}

Entropy regularization is evaluated along every sampled trajectory.
Using the path-conditional distributions defined above, the
objective sums, rather than averages, their entropies across the
complete construction:
\begin{align}
    \mathcal{H}(p)
      &=-\sum_j p_j\log p_j,
      \label{eq:categorical_entropy}\\
    H_b^{(m)}
      &=\mathcal{H}(p_{b,0})
        +\sum_{t=1}^{\numnodes-1}
          \mathcal{H}(p_{b,t}^{(m)}),
        \label{eq:path_entropy}\\
    \bar H
      &=\frac{1}{B\numcandidates}
        \sum_{b,m}H_b^{(m)}.
    \label{eq:mean_path_entropy}
\end{align}
For optimizer step $t\in\{0,\ldots,T-1\}$, its coefficient is
linearly annealed,
\begin{equation}
    \entropycoef_t
    =\entropycoef_0
     +(\entropycoef_T-\entropycoef_0)\frac{t}{T-1}.
    \label{eq:entropy_schedule}
\end{equation}
The full \mcsrl{} objective is
\begin{equation}
    \mathcal{L}_{\mathrm{MCS\text{-}RL}}
    =\mathcal{L}_{\mathrm{PG}}
     +\guidancecoef\mathcal{L}_{\mathrm{lead}}
     -\entropycoef_t\bar H.
    \label{eq:mcs_full_objective}
\end{equation}
We use $\numcandidates=3$, $\guidancecoef=0.5$,
$\entropycoef_0=0.0125$, and $\entropycoef_T=0$.  A single neural
forward pass supplies all candidates, which are then sampled from the
shared start probabilities and edge logits.  Consequently,
\mcsrl{} adds no parameters and leaves inference unchanged.

\paragraph{Single-candidate reference objective.}
The PG reference used by \narR{} and \grnPG{} samples one tour and
uses greedy decoding as its baseline.  If
$D_b=C_b^{\mathrm{s}}-C_b^{\mathrm{g}}$ and
$\bar D=B^{-1}\sum_b D_b$, we use the batch-centered loss
\begin{equation}
    \mathcal{L}_{\mathrm{single}}
    =\frac{1}{B}\sum_{b=1}^{B}
      (D_b-\bar D)\ell_b^{\mathrm{s}}.
    \label{eq:single_pg}
\end{equation}
This distinction is important for the architecture--training
comparison reported in Section~\ref{sec:experiments}.

\section{Experiments}
\label{sec:experiments}

\begin{table*}[!t]
    \centering
    \caption{Beam-1000 mean gap relative to the Concorde reference (\%) for the single-seed $2\!\times\!2$ comparison. Lower is better. \tspfifty\ is the validation set used for checkpoint selection; \tsponehundred\ is a correlated size diagnostic; \tsplib\ uses normalized model inputs and original EUC\_2D matrices for cost evaluation.}
    \label{tab:main_2x2}
    \small
    \setlength{\tabcolsep}{6pt}
    \begin{tabular}{llccc}
        \toprule
        Encoder & Objective & \tspfifty\ validation & \tsponehundred\ diagnostic & \tsplib\ (27) \\
        \midrule
        \multirow{2}{*}{\nar} & PG & 0.42 & 2.73 & 17.12 \\
        & \mcsrl & 0.36 & 2.17 & 21.98 \\
        \midrule
        \multirow{2}{*}{Geometry-aware} & PG & 0.37 & 1.54 & 5.41 \\
        & \mcsrl & \textbf{0.32} & \textbf{1.26} & \textbf{3.60} \\
        \bottomrule
    \end{tabular}
\end{table*}

\begin{figure*}[!t]
    \centering
    \includegraphics[width=0.94\textwidth]{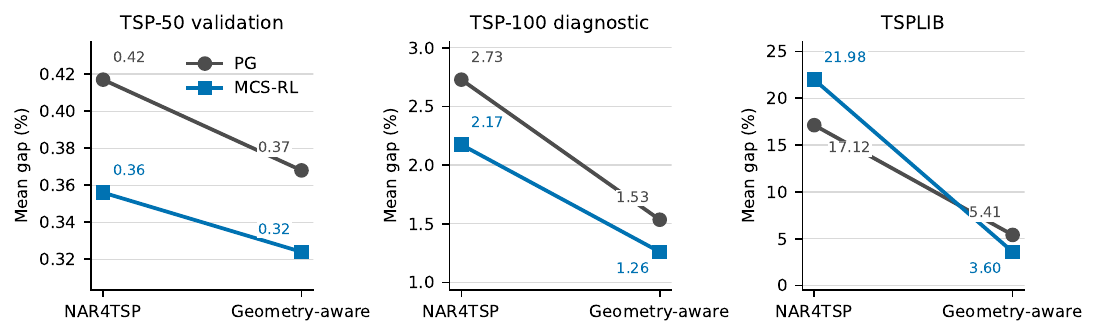}
    \caption{Encoder and training-objective comparison under Beam-1000 decoding. Each point is the aggregate result from one trained model (single seed). On \tsplib, the direction of the \mcsrl\ difference reverses between encoders; no confidence intervals are available.}
    \label{fig:interaction}
\end{figure*}

\subsection{Protocol and Claim Scope}

\paragraph{Training and model selection.}
All neural variants are trained online on uniformly sampled \tspfifty\ instances in $[0,1]^2$.
The common configuration uses hidden dimension 128, eight attention heads, six encoder layers, a two-layer edge decoder, $k$-NN graphs with $k=10$, and batch size 64.
Optimization uses Adam~\cite{kingma2015adam} with learning rate $10^{-4}$ and no weight decay for 1,000 epochs of 2,500 steps.
The single reported training seed is 1234.
For \mcsrl, the number of sampled candidates is $\numcandidates=3$, the winner-guidance coefficient is $\guidancecoef=0.5$, and the entropy coefficient decreases linearly from $0.0125$ to zero.

The fixed \tspfifty\ validation set contains 10,000 instances generated with pseudorandom seed 1234.
We evaluate this set after every epoch and select the checkpoint with the lowest greedy mean tour length.
Because the set participates in model selection, all reported \tspfifty\ entries, including Beam-1000 results for the selected checkpoint, are validation rather than independent test results.

\paragraph{Scale and distribution diagnostics.}
The \tsponehundred\ set also contains 10,000 instances and was generated from the same pseudorandom sequence and seed as the \tspfifty\ validation set.
The coordinate sequence used for \tspfifty\ is an exact prefix of that used for \tsponehundred; regrouping these values into 50- and 100-node instances induces dependence between the two sets.
We therefore use \tsponehundred\ only as a \emph{correlated size-extrapolation diagnostic}, not as an independent test set.

Cross-distribution behavior is assessed on 27 \tsplib\ EUC\_2D instances~\cite{reinelt1991tsplib} with 51--299 nodes.
For neural inference, each instance is normalized to $[0,1]^2$; tour costs and gaps relative to the Concorde reference are computed with the original integer EUC\_2D distance matrix.
For size-resolved reporting, the instances are grouped into four node ranges: 51--76 (5 instances), 99--101 (8), 105--150 (7), and 159--299 (7).
No \tsplib\ instance is used for checkpoint selection.

\paragraph{Variants, decoding, and metrics.}
The evaluation forms a $2\!\times\!2$ design: the reproduced \nar\ encoder or the geometry-aware encoder is paired with either single-candidate policy gradient (PG) or \mcsrl.
The resulting configurations are \narR, \narMCS, \grnPG, and \grnMCS; the last is the full \method\ model.
All variants use the same Greedy, Beam-100, and Beam-1000 decoders.
For $N$ instances, we report the mean per-instance gap
\begin{equation}
    \mathrm{Gap}(\%)=\frac{100}{N}\sum_{i=1}^{N}
    \frac{C_i-C_i^{\mathrm{ref}}}{C_i^{\mathrm{ref}}},
    \label{eq:eval_gap}
\end{equation}
where $C_i^{\mathrm{ref}}$ is obtained with Concorde~\cite{applegate2006tsp}.
For the random sets, Euclidean distances are multiplied by $10^6$ and rounded to integers for Concorde; the returned tour is then rescored on the original floating-point coordinates.
The reported values are aggregate results from one training seed; they do not quantify training-seed variability, and no significance tests are performed.

Experiments were conducted using one Intel Xeon Silver 4214R CPU and one NVIDIA GeForce RTX 3080 Ti GPU with Python 3.12.3, PyTorch 2.7.0, and CUDA 12.8.
Random-set neural evaluation uses a nominal batch size of 128, while \tsplib\ uses batch size 1.
Additional decoder and timing details appear in Appendix~\ref{app:experimental_detail}.

\subsection{\texorpdfstring{Single-Seed $2\!\times\!2$ Comparison}{Single-Seed 2x2 Comparison}}

Table~\ref{tab:main_2x2} reports the common Beam-1000 comparison.
On the \tspfifty\ validation set, all three alternative configurations have lower reported gaps than \narR, although the differences are small and are measured on the set used for checkpoint selection.
On the correlated \tsponehundred\ diagnostic, replacing the reproduced encoder while retaining PG changes the gap from $2.73\%$ to $1.54\%$; adding \mcsrl\ to the geometry-aware encoder changes it further to $1.26\%$.

The largest contrasts occur on \tsplib.
With PG fixed, the geometry-aware encoder changes the overall gap from $17.12\%$ to $5.41\%$.
With the geometry-aware encoder, changing PG to \mcsrl\ changes the gap from $5.41\%$ to $3.60\%$; with the reproduced encoder, it changes the gap from $17.12\%$ to $21.98\%$.
The \tsplib\ point estimates therefore show an encoder-dependent pattern; they do not support an encoder-independent benefit of multi-candidate training.

Figure~\ref{fig:interaction} displays the same values as an interaction plot.
For both random-instance sets, the \mcsrl\ line lies below the PG line for both encoders.
On \tsplib, however, the lines cross: the reported \mcsrl\ point has a higher gap with the reproduced encoder and a lower gap with the geometry-aware encoder.
Each cell contains one trained model, and the encoder change bundles several architectural components; this pattern supports only a descriptive comparison, not component-level causal attribution.

\subsection{Decoder Behavior}

Wider beam search lowers the gap for every neural variant on all three evaluation sets (Appendix~\ref{app:decode_results}).
For \grnMCS, the sequence from Greedy to Beam-100 to Beam-1000 is $0.95\%\!\rightarrow\!0.43\%\!\rightarrow\!0.32\%$ on the \tspfifty\ validation set, $2.34\%\!\rightarrow\!1.52\%\!\rightarrow\!1.26\%$ on the correlated \tsponehundred\ diagnostic, and $5.27\%\!\rightarrow\!3.96\%\!\rightarrow\!3.60\%$ over \tsplib.
The smaller Beam-100-to-Beam-1000 change is consistent with diminishing returns for this decoder configuration, but the available aggregates do not identify whether the pattern arises from score calibration, beam diversity, or instance composition.

\section{Limitations and Validity}
\label{sec:limitations}

\paragraph{Evaluation design.}
The reported study uses one training seed for each variant.
The fixed \tspfifty\ set is used both for checkpoint selection and for the corresponding table entry, so that entry measures selected validation performance rather than held-out generalization.
The \tsponehundred\ set reuses the initial coordinate sequence of the \tspfifty\ set and is therefore a correlated size diagnostic.
An independently generated random test suite is required to estimate generalization to new samples from the random Euclidean distribution.
The 27 \tsplib\ instances were not used for selection, but they form a small, heterogeneous benchmark; their grouped means should not be read as population estimates.

\paragraph{Statistical and attribution limits.}
Without multiple independent training seeds or consistent per-instance results, the present study cannot report training-seed variance, paired confidence intervals, or significance tests.
All numerical comparisons are descriptive single-seed aggregate results.
Moreover, the encoder factor in the $2\!\times\!2$ design changes centered features, radial bases, attention, edge updates, normalization, feed-forward blocks, cross-layer mixing, and parameter count (0.912M versus 2.024M) together.
The design therefore supports comparison of the complete encoder bundles, not attribution to any one component or to model capacity separately.
The candidate count, entropy schedule, winner coefficient, RBF count, and graph neighborhood size were fixed rather than systematically swept.

\paragraph{Input and problem scope.}
The \tsplib\ coordinate tensors are normalized to $[0,1]^2$ before neural inference, while original EUC\_2D matrices define tour cost.
Consequently, the experiment probes changes in node count and normalized spatial layout, not robustness to absolute coordinate scale.
The study covers symmetric Euclidean TSP only, fixes $k=10$, and uses no local-search post-processing such as 2-opt.
The conclusions do not directly extend to asymmetric, non-Euclidean, or constrained routing problems.

\paragraph{Efficiency and reproducibility.}
Neural timings use GPU inference whereas Concorde and \lkh\ use CPU per-instance processes.
The reported latency probe repeats only the first instance after one warmup; full-set runtime is more representative, but remains hardware- and implementation-specific.
These measurements describe throughput in the specified hardware and software setting and do not establish hardware-independent algorithmic speedups.
Architecture and optimization details were verified against the implementation used for the original experiments.
Numerical tables were reconstructed from the available dataset- and group-level summaries rather than from consistent per-instance records.
Raw timing observations were also unavailable, and the missing measurements were not regenerated.

\section{Conclusion}
\label{sec:conclusion}

We studied two interventions for non-autoregressive Euclidean TSP solving: an encoder that exposes relative position and pairwise distance more directly, and a reinforcement-learning objective that compares multiple tours sampled for the same instance.
\method\ augments node and edge representations with centered geometric features, learnable radial distance bases, distance-aware attention, explicit edge updates, and cross-layer mixing.
\mcsrl\ combines a leave-one-out adaptive baseline with winner-candidate guidance and an annealed entropy bonus.

The $2\!\times\!2$ comparison gives a more specific conclusion than the full model result alone.
Under PG, the geometry-aware encoder has lower gaps than the reproduced encoder on the correlated \tsponehundred\ diagnostic and \tsplib.
With the geometry-aware encoder, \mcsrl\ is associated with lower gaps; with the reproduced encoder, it is associated with a higher \tsplib\ gap.
The single-seed comparison therefore does not establish an encoder-independent benefit from multi-candidate comparison; it documents an encoder-dependent pattern that requires replication.

These findings are empirical and limited to one training seed, Euclidean instances, and fixed decoder configurations.
Multi-seed training, an evaluation set separated from checkpoint selection, and experiments that isolate size shift from spatial-distribution shift are needed before drawing broader conclusions.
Within that scope, the results motivate multi-seed and independently held-out evaluation of the complete geometry-aware encoder bundle, with multi-candidate training evaluated across encoder choices.

\bibliographystyle{plainnat}
{\small\bibliography{refs}}

\appendix
\section{Additional Results}
\label{app:experimental_detail}

This appendix reports the available aggregate results for all decoder settings.
As in the main text, \tspfifty\ is a validation set used for checkpoint selection and \tsponehundred\ is a correlated size-extrapolation diagnostic; neither is treated as an independent test set.
Each cell in the $2\!\times\!2$ design represents one trained model.
Because consistent per-instance results are unavailable, we report only dataset- and group-level aggregates and omit per-instance tours.

\subsection{Random-Set Decoding}
\label{app:decode_results}

Table~\ref{tab:appendix_random_decode} reports results for all three decoding strategies.
The Concorde and \lkh\ rows provide solver references; their values are not neural decoding results.
The gap is the mean of per-instance relative gaps, not the relative difference between the two displayed mean lengths.

\begin{table*}[t]
    \centering
    \captionsetup{width=0.90\textwidth,justification=raggedright,singlelinecheck=false}
    \caption{Complete random-set aggregate results. \tspfifty\ is the validation set used for checkpoint selection and \tsponehundred\ is the correlated size diagnostic. Length is mean tour length; Gap is the mean per-instance gap relative to the Concorde reference.}
    \label{tab:appendix_random_decode}
    \small
    \setlength{\tabcolsep}{5pt}
    \begin{tabular*}{0.90\textwidth}{@{\extracolsep{\fill}}llrrrr@{}}
        \toprule
        Method & Decode & \multicolumn{2}{c}{\tspfifty\ validation} & \multicolumn{2}{c}{\tsponehundred\ correlated diagnostic} \\
        \cmidrule(lr){3-4}\cmidrule(lr){5-6}
        & & Length & Gap (\%) & Length & Gap (\%) \\
        \midrule
        Concorde & --- & 5.6948 & 0.000 & 7.7641 & 0.000 \\
        \lkh & --- & 5.6949 & 0.001 & 7.7644 & 0.003 \\
        \midrule
        \multirow{3}{*}{\narR} & Greedy & 5.7817 & 1.520 & 8.1439 & 4.888 \\
        & Beam-100 & 5.7306 & 0.626 & 8.0224 & 3.325 \\
        & Beam-1000 & 5.7187 & 0.417 & 7.9761 & 2.728 \\
        \midrule
        \multirow{3}{*}{\narMCS} & Greedy & 5.7690 & 1.298 & 8.0884 & 4.175 \\
        & Beam-100 & 5.7244 & 0.517 & 7.9742 & 2.704 \\
        & Beam-1000 & 5.7152 & 0.356 & 7.9330 & 2.174 \\
        \midrule
        \multirow{3}{*}{\grnPG} & Greedy & 5.7605 & 1.149 & 7.9789 & 2.766 \\
        & Beam-100 & 5.7239 & 0.509 & 7.9086 & 1.860 \\
        & Beam-1000 & 5.7159 & 0.368 & 7.8833 & 1.535 \\
        \midrule
        \multirow{3}{*}{\grnMCS} & Greedy & 5.7491 & 0.948 & 7.9457 & 2.337 \\
        & Beam-100 & 5.7197 & 0.435 & 7.8819 & 1.517 \\
        & Beam-1000 & \textbf{5.7134} & \textbf{0.324} & \textbf{7.8620} & \textbf{1.261} \\
        \bottomrule
    \end{tabular*}
\end{table*}

\subsection{\tsplib\ Group Results}

Table~\ref{tab:appendix_tsplib_groups} reports the available group-level results for all three decoding strategies.
The four columns group instances by node count: 51--76 (5 instances), 99--101 (8), 105--150 (7), and 159--299 (7).
Overall is the mean over all 27 per-instance gaps and therefore weights groups by their instance counts.
All costs use the original EUC\_2D matrices.

\begin{table*}[t]
    \centering
    \captionsetup{width=0.90\textwidth,justification=raggedright,singlelinecheck=false}
    \caption{Complete \tsplib\ group-level mean gap (\%). Values are rounded to two decimals; lower is better.}
    \label{tab:appendix_tsplib_groups}
    \small
    \setlength{\tabcolsep}{5pt}
    \begin{tabular*}{0.90\textwidth}{@{\extracolsep{\fill}}llrrrrr@{}}
        \toprule
        Method & Decode & 51--76 & 99--101 & 105--150 & 159--299 & Overall \\
        \midrule
        Concorde & --- & 0.00 & 0.00 & 0.00 & 0.00 & 0.00 \\
        \lkh & --- & 0.00 & 0.00 & 0.00 & 0.00 & 0.00 \\
        \midrule
        \multirow{3}{*}{\narR} & Greedy & 21.46 & 24.89 & 21.14 & 24.96 & 23.30 \\
        & Beam-100 & 14.26 & 18.10 & 17.51 & 21.70 & 18.17 \\
        & Beam-1000 & 13.27 & 16.82 & 16.43 & 20.91 & 17.12 \\
        \midrule
        \multirow{3}{*}{\narMCS} & Greedy & 27.39 & 28.66 & 28.31 & 29.08 & 28.44 \\
        & Beam-100 & 22.83 & 22.30 & 22.91 & 25.28 & 23.33 \\
        & Beam-1000 & 21.06 & 20.74 & 21.31 & 24.72 & 21.98 \\
        \midrule
        \multirow{3}{*}{\grnPG} & Greedy & 3.16 & 2.28 & 5.64 & 16.77 & 7.07 \\
        & Beam-100 & 1.97 & 1.35 & 4.33 & 15.45 & 5.89 \\
        & Beam-1000 & 1.75 & 0.99 & 3.86 & 14.62 & 5.41 \\
        \midrule
        \multirow{3}{*}{\grnMCS} & Greedy & 3.12 & 2.34 & 5.47 & 9.94 & 5.27 \\
        & Beam-100 & 1.53 & 1.17 & 3.77 & 9.07 & 3.96 \\
        & Beam-1000 & \textbf{1.30} & \textbf{0.89} & \textbf{3.54} & \textbf{8.41} & \textbf{3.60} \\
        \bottomrule
    \end{tabular*}
\end{table*}

\subsection{Efficiency Measurements}

Table~\ref{tab:appendix_efficiency} reports efficiency measurements for the four neural configurations in the main comparison and the two classical solvers.
Evaluation of the random-instance sets uses a nominal GPU batch size of 128, while \tsplib\ uses batch size 1.
Traditional solvers run on CPU, one instance per process; \lkh\ is executed once per instance.
``Probe latency'' is the mean of three repetitions on the first instance after one warmup, whereas total runtime covers the full set.
Because the latency probe and full-set runtime use different batching conditions, they are not directly convertible.

\begin{table*}[t]
    \centering
    \captionsetup{width=0.90\textwidth,justification=raggedright,singlelinecheck=false}
    \caption{Descriptive efficiency under the specified heterogeneous CPU/GPU protocol. Neural rows use Beam-1000. Probe latency is not a dataset-wide average. No uncertainty estimates are available.}
    \label{tab:appendix_efficiency}
    \small
    \setlength{\tabcolsep}{5pt}
    \begin{tabular*}{0.90\textwidth}{@{\extracolsep{\fill}}llrrr@{}}
        \toprule
        Dataset & Method & Probe latency (ms) & Total runtime (s) & Throughput (inst./s) \\
        \midrule
        \multirow{6}{*}{\tspfifty\ validation}
        & Concorde & 96.00 & 829.05 & 12.1 \\
        & \lkh & 63.60 & 632.58 & 15.8 \\
        & \narR & 42.12 & 10.10 & 989.8 \\
        & \narMCS & 44.17 & 10.13 & 987.4 \\
        & \grnPG & 45.25 & 9.31 & 1074.7 \\
        & \grnMCS & 33.16 & 9.22 & 1084.8 \\
        \midrule
        \multirow{6}{*}{\tsponehundred\ diagnostic}
        & Concorde & 407.20 & 2824.38 & 3.5 \\
        & \lkh & 114.66 & 1857.20 & 5.4 \\
        & \narR & 54.45 & 36.72 & 272.3 \\
        & \narMCS & 58.40 & 36.93 & 270.8 \\
        & \grnPG & 57.97 & 33.00 & 303.0 \\
        & \grnMCS & 57.91 & 33.01 & 303.0 \\
        \midrule
        \multirow{6}{*}{\tsplib\ (27)}
        & Concorde & 642.9 & 17.58 & 1.54 \\
        & \lkh & 1250.4 & 37.21 & 0.73 \\
        & \narR & 79.5 & 2.20 & 12.27 \\
        & \narMCS & 79.4 & 2.17 & 12.44 \\
        & \grnPG & 81.4 & 2.29 & 11.79 \\
        & \grnMCS & 83.5 & 2.28 & 11.84 \\
        \bottomrule
    \end{tabular*}
\end{table*}

The geometry-aware variants contain 2.024M parameters, compared with 0.912M for the reproduced \nar\ variants.
Peak GPU allocation for the geometry-aware versus reproduced variants is 1.251 versus 1.543\,GB on \tspfifty, and 4.877 versus 6.097\,GB on \tsponehundred.
This difference reflects implementation choices, including the avoidance of explicit pairwise tensor replication, and should not be interpreted as an asymptotic complexity difference.

\end{document}